\documentclass{article}

\usepackage{microtype}
\usepackage{graphicx}
\usepackage{subfigure}
\usepackage{booktabs} % for professional tables

\usepackage{hyperref}

\usepackage[accepted]{icml2025}

\usepackage{amsmath}
\usepackage{amssymb}
\usepackage{mathtools}
\usepackage{amsthm}

\usepackage[capitalize,noabbrev]{cleveref}

\usepackage{comment}
\usepackage{paralist}

\usepackage{xr}
\usepackage[algo2e, ruled, lined, linesnumbered, commentsnumbered, longend]{algorithm2e}

\usepackage{multirow}
\usepackage{makecell}
\usepackage{paralist}
\usepackage{threeparttable} % tablenotes

\usepackage{xcolor}         % colors
\usepackage{color, colortbl}

\usepackage{dsfont}
\usepackage{bbding}
\usepackage{pifont}

\newcommand{\first}[1]{\textbf{#1}}
\newcommand{\second}[1]{\underline{#1}}

\newcommand{\set}[1]{\{ #1 \}}

\newcommand{\dataset}[1]{\mathcal{D}_{#1}}
\newcommand{\category}[1]{\mathcal{C}_{#1}}
\newcommand{\task}[1]{\mathcal{T}_{#1}}
\newcommand{\norm}[1]{\lVert{#1}\rVert}

\newcommand{\loss}{\mathcal{L}}
\newcommand{\real}{\mathbb{R}}
\newcommand{\prob}{\mathcal{P}}
\newcommand{\expect}{\mathbb{E}}
\newcommand{\kldiv}{\mathcal{D}_{KL}}

\newcommand{\gmmdist}{\mathcal{N}}

\newcommand{\proposed}{ViRN}

\theoremstyle{plain}

\theoremstyle{definition}

\theoremstyle{remark}

\usepackage[textsize=tiny]{todonotes}

\icmltitlerunning{Learning and Forgetting from a Bayesian Perspective}

\begin{document}

\twocolumn[
\icmltitle{ViRN: Variational Inference and Distribution Trilateration \\ for Long-Tailed Continual Representation Learning}



\icmlsetsymbol{equal}{*}

\begin{icmlauthorlist}
\icmlauthor{Hao Dai}{yyy,soton}
\icmlauthor{Chong Tang}{yyy,soton}
\icmlauthor{Jagmohan Chauhan}{yyy,soton}
\end{icmlauthorlist}

\icmlaffiliation{yyy}{Department of Computer Science, UCL Centre for Artificial Intelligence, University College London, London, UK}
\icmlaffiliation{soton}{University of Southampton, Southampton, UK}

\icmlcorrespondingauthor{Hao Dai}{daihaovigg@gmail.com}

\icmlkeywords{Continual Learning, Variational AutoEncoder, Bayesian Inference}

\vskip 0.3in
]



\printAffiliationsAndNotice{}  % leave blank if no need to mention equal contribution

\begin{abstract}
  Continual learning (CL) with long-tailed data distributions remains a critical challenge for real-world AI systems, where models must sequentially adapt to new classes while retaining knowledge of old ones—despite severe class imbalance. Existing methods struggle to balance stability and plasticity, often collapsing under extreme sample scarcity. To address this, we propose ViRN, a novel CL framework that integrates variational inference (VI) with distributional trilateration for robust long-tailed learning. First, we model class-conditional distributions via a Variational Autoencoder to mitigate bias toward head classes. Second, we reconstruct tail-class distributions via Wasserstein distance-based neighborhood retrieval and geometric fusion, enabling sample-efficient alignment of tail-class representations. Evaluated on six long-tailed classification benchmarks—including speech (e.g., rare acoustic events, accents) and image tasks—ViRN achieves a $10.24\%$ average accuracy gain over state-of-the-art methods.
\end{abstract}

\section{Introduction}

Continual learning (CL)~\cite{10444954} enables models to incrementally acquire knowledge from evolving data streams, akin to human learning.
Class-incremental learning (CIL)~\cite{Rebuffi2016iCaRLIC, NEURIPS2023_15294ba2, 10203164} specifically tackles sequential learning of new classes while preserving historical performance without data revisitation - crucial for real-world applications like adaptive voice assistants and intelligent surveillance. 
While existing CIL methods achieve promising results in controlled settings, their real-world deployment remains hindered by non-stationary, task-agnostic data streams—particularly under long-tailed distributions, where tail classes with sparse samples suffer catastrophic forgetting and bias amplification~\cite{Kirkpatrick2016OvercomingCF, 8107520}.

The core challenge in long-tailed CIL (LCIL)~\cite{10203164, He2024GradientRT} lies in extreme class imbalance: abundant head-class samples dominate training while tail classes face severe underrepresentation (e.g., rare acoustic events like glass breaking, underrepresented speech accents, or infrequent objects such as endangered species). Traditional approaches using replay buffers or regularization fail to balance knowledge retention and unbiased representation learning~\cite{10.5555/3692070.3693902}. Recent pre-trained models (PTMs)~\cite{ijcai2024p0924} offer promise through rich, transferable representations from large-scale audio or visual corpora that enhance generalization and mitigates forgetting, while generative classifiers can decouple learning from data bias by synthesizing tail-class features~\cite{NEURIPS2023_15294ba2} via VAEs~\cite{kingma2022autoencodingvariationalbayes}/diffusion models~\cite{pmlr-v37-sohl-dickstein15}. Yet, these methods struggle with extreme data scarcity: limited samples lead to inaccurate distribution estimation, as generative models cannot reliably capture true class-conditional patterns from minimal observations.

Motivated by this limitation, we propose to reconstruct tail-class distributions by leveraging geometric relationships in the PTM embedding space, where semantically similar classes (e.g., acoustically proximate speech accents) cluster together.
Our design first employs Variational AutoEncoder (VAE) to estimate initial class distributions, then reconstructs tail-class representations via Wasserstein distance-based trilateration using nearest neighbors, and finally fuses these estimates adaptively to refine robustness. Our contributions are:
1) A variational inference framework for modeling class-conditional distributions in long-tailed CIL, mitigating bias toward head classes.
2) A novel LCIL framework, {\proposed}, leveraging a trilateration-based distribution alignment method that reconstructs tail-class representations using Wasserstein neighbors and fuses them with VI estimates.
3) Extensive experiments on six long-tailed acoustic and image benchmarks demonstrate a $10.24\%$ average accuracy gain over state-of-the-art methods, with notable improvements in speech-related tasks.

\section{Related Work}\label{sec:related}

Conventional CIL methods combat catastrophic forgetting through regularization (e.g., EWC~\cite{Kirkpatrick2016OvercomingCF}, LwF~\cite{8107520}), dynamic architectures~\cite{Yoon2017LifelongLW}, or replay buffers~\cite{Rebuffi2016iCaRLIC}. However, these approaches struggle with long-tailed data: replay exacerbates head-class bias, while regularization ignores inter-phase imbalance. 
Conversely, long-tailed techniques—such as re-weighting~\cite{8953804}, margin adjustment~\cite{10.5555/3454287.3454427}, or synthetic generation~\cite{Liu2020DeepRL}—assume static data and lack incremental adaptation. Recent hybrid attempts (e.g., imbalance-aware replay~\cite{9878763}) remain limited by raw data dependency.
The advent of pre-trained models (PTMs) has transformed both fields. Vision/audio PTMs (e.g., Wav2Vec2~\cite{NEURIPS2020_92d1e1eb}, HuBERT~\cite{10.1109/TASLP.2021.3122291}, CLAP~\cite{Wu2022LargeScaleCL}) enable rehearsal-free CIL via prompt tuning~\cite{9878681} or distillation~\cite{10.1007/978-3-030-58565-5_6}, while generative classifiers (Gaussian prototypes~\cite{NEURIPS2023_15294ba2}, diffusion synthesis~\cite{shao2024diffult}) rebalance predictions by modeling class distributions in PTM spaces. 
Yet, these approaches struggle to estimate accurate distributions for tail classes with extremely sparse samples, especially in incremental deployments where new tail classes arrive sequentially.
Our work bridges this gap through variational inference-driven distributional trilateration, enabling sample-efficient long-tailed CIL without raw data replay.

\section{Methodology}\label{sec:method}

\subsection{Problem Formulation}
\label{sec:problem}
Given a long-tailed dataset $\dataset{}$ containing $\category{}$ classes, it is partitioned into $\task{}$ sequential tasks, with each task introducing a subset of new classes. 
Long-tailed class incremental learning (LCIL) aims to address the sequential training of models where: (1) class distributions exhibit a heavy long-tailed skew, and (2) tasks arrive incrementally with disjoint class sets.
At each task $\task{t}$, the number of samples 
$n_i$ for class $c_i$ follows a power-law distribution as follow:
\begin{equation}
  \label{eq:problem}
\setlength\abovedisplayskip{2pt}
  \setlength\belowdisplayskip{2pt}
  %\footnotesize
  \begin{aligned}
    n_i = \max (1, \tilde{n_i} \cdot \rho^{\frac{c_i}{|\category{}| - 1}}), \quad \forall c_i \in \category{}
  \end{aligned}
\end{equation}
, where $\tilde{n_i}$ is the number of samples of class $i$ in the whole set, $\rho \in (0, 1]$ controls imbalance severity (e.g., smaller $\rho$  yields extreme skew). 
 Over $\task{}$ tasks, the model $f_{\theta} : \mathcal{X} \rightarrow \mathcal{Y}$ observes disjoint class sets $\set{\category{t} \mid t = 1 \ldots, \task{}}$ and data $\set{\dataset{t} = (x_i^t,y_i^t)}$, where $y^t_i \in \category{t} = \set{c_1, \ldots, c_t}$.
 The objective is to incrementally adapt $f_{\theta}$ to new tasks $\task{t}$ without accessing prior data $\dataset{1:t-1}$, while maintaining balanced performance across all seen classes and mitigating catastrophic forgetting.

%LCIL faces two intertwined challenges: distributional imbalance and incremental adaptation:
%1) Distributional Imbalance, the long-tailed nature of each task induces intra-task bias, where models overfit to head classes while underfitting tail classes. 
%2) Incremental Adaptation, the stability-plasticity dilemma emerges—adapting to new tasks leads to catastrophic forgetting of prior knowledge, whereas resisting forgetting hampers learning new tail classes. 

\subsection{Variational AutoEncoder based Nearest Neighbor Classifier}

Through self-supervised pre-trained models, raw features can be mapped to a latent space, and we further encode this latent space to achieve compact and continual representation learning. To validate the effectiveness of the learned representations, we adopt a generative classifier as the downstream evaluation task, which directly models class-conditional feature distributions \(\prob(x_i | y_i)\), circumventing the bias-prone discriminative paradigm. 
A canonical implementation is prototypical learning~\cite{NIPS2017_cb8da676}, which estimates a class prototype \(f_c \in \real^M\) as the feature mean.
Classification follows the nearest class mean rule:
\begin{equation}
  \label{eq:ncm}
  \setlength\abovedisplayskip{2pt}
  \setlength\belowdisplayskip{2pt}
  %\footnotesize
  \prob(y_i = c | x_i) = \frac{\exp(-d(x_i, c))}{\sum_{f_c^{\prime}} \exp(-d(x_i, f_c^{\prime}))}
\end{equation}
where $d(\cdot)$ is a distance metric. 
However, in high-dimensional spaces, Euclidean distances suffer from hyperspherical concentration, leading to degraded discrimination as feature dimensions grow.
To mitigate this, we model each class as a Gaussian distribution \(\gmmdist(\mu_c, \Sigma_c)\), estimated via:
\begin{equation}
\setlength\abovedisplayskip{2pt}
  \setlength\belowdisplayskip{2pt}
  %\footnotesize
  \label{eq:gmm}
    \mu_c = \frac{1}{n_c}\sum_{i}^{n_c} x_i, \quad \Sigma_c = \frac{1}{n_c}\sum_{i}^{n_c} (x_i - \mu_c)(x_i - \mu_c)^T 
\end{equation}
generalizing the distance to the Mahalanobis similarity:
\begin{equation}
\setlength\abovedisplayskip{2pt}
  \setlength\belowdisplayskip{2pt}
  \label{eq:sim}
  %\footnotesize
  d(x_i, c) = (x_i - \mu_c)^T \Sigma_c^{-1} (x_i - \mu_c)
\end{equation}
This adaptively scales distances using class-specific covariance, alleviating the curse of dimensionality.

While Gaussian modeling improves high-dimensional discrimination, tail classes with sparse samples ($n_c \ll M$) render $\Sigma_c$ estimation numerically unstable. To address this, we treat $\mu_c$ and $\Sigma_c$ as latent variables $z$ and approximate their posterior $p(z|x)$ using VAE.
We optimize the Evidence Lower Bound (ELBO)~\cite{kingma2022autoencodingvariationalbayes}:
\begin{equation}
  \setlength\abovedisplayskip{3pt}
  \setlength\belowdisplayskip{3pt}
  %\footnotesize
  \label{eq:elbo}
    \loss_{ELBO} = \expect_{q(z|x)}[\ln p(x|z)] - \kldiv(q(z|x) || p(z))
\end{equation}
where $p(z)= \gmmdist(0,\mathbb{I})$ acts as a conjugate prior, enabling closed-form updates. By learning $q(z|x)$ via variational inference (VI)~\cite{Hoffman2012StochasticVI}, we stabilize distribution estimation for tail classes, even with few samples.

\subsection{Neighborhood-Aware Distribution Refinement}
While VI provides robust posterior estimates, it fails to account for distributional shifts induced by long-tailed data, where sparse tail-class samples lead to biased or overconfident estimates. To address this, we propose a redistribution mechanism that leverages geometric relationships between classes in the embedding space to refine the VI-estimated distributions.

\begin{figure}[htb]
\vspace{-3mm}
\centering
\includegraphics[width=0.8\linewidth]{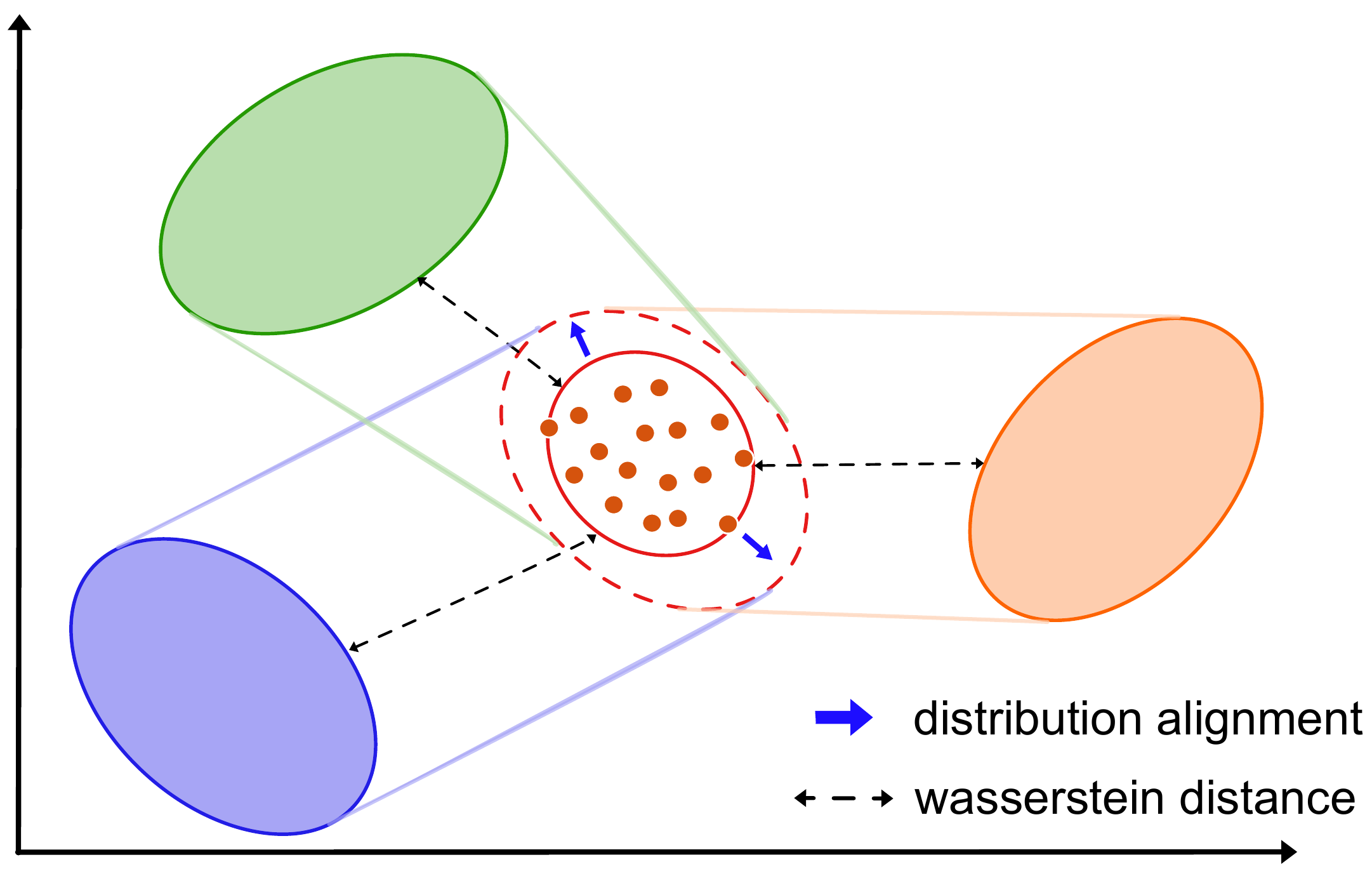}
\vspace{-2mm}
\caption{The trilateration method for reconstructing distributions.}
\label{fig:redist}
\vspace{-5mm}
\end{figure}

To identify semantically similar classes, we compute the 2-Wasserstein distance \(W_2\) between pairs of Gaussian distributions \(\gmmdist(\mu_i, \Sigma_i)\) and \(\gmmdist(\mu_j, \Sigma_j)\). For Gaussians, $W_2$ admits a closed-form solution:
\begin{equation}
  \label{eq:wdist}
  \setlength\abovedisplayskip{3pt}
  \setlength\belowdisplayskip{3pt}
  %\footnotesize
  \begin{aligned}
    W^2_2(i, j) & = \norm{\mu_{i} - \mu_{j}}^2 \\
                & + \text{Tr}(\Sigma_{i} + \Sigma_{j} - 2(\Sigma_{j}^{1/2}\Sigma_{i}\Sigma_{j}^{1/2})^{1/2} )
\end{aligned}
\end{equation}
which captures both mean displacement and covariance misalignment.
Classes with minimal $W_2$ distances are retained as nearest neighbors, forming a graph where edges encode distributional similarity.
Consequently,  for a tail class $i$, its $K$ nearest head classes $N_i$ based on $W_2$ distance is computed as follows:
\begin{equation}
  \label{eq:neighbor}
  \setlength\abovedisplayskip{3pt}
  \setlength\belowdisplayskip{3pt}
  %\footnotesize
  \begin{aligned}
N_i = \operatorname{argmin}_{\substack{S \subseteq \mathcal{C},\\ |S| = k}} \sum_{j \in S} W_2(i, j)
  \end{aligned}
\end{equation}
We use the $K$-nearest neighbors to reconstruct the tail class distribution via geometric interpolation.
Through the trilateration method, the target class's distribution can be reconstructed, as shown in Figure~\ref{fig:redist}.
This provides an alternative perspective to estimate the distribution of the tail class, serving as a reference for adjusting the initial estimation.
%By fusing the initial distribution with the reconstructed one, it effectively borrows the statistical support from neighbors while preserving the covariance structure.

For this, we define the weight of each neighbor for distribution reconstruction, derived from the $W_2$ distance as follows:
\begin{equation}
  \label{eq:weight}
\setlength\abovedisplayskip{2pt}
  \setlength\belowdisplayskip{2pt}
  %\footnotesize
  \begin{aligned}
    \omega_{j} = \frac{1/W_2(i, j)}{\sum^k_{m \in N_i}1/W_2(i, m)}
  \end{aligned}
\end{equation}
Additionally, due to the significant difference in the number of samples between head and tail classes, we design an adaptive dynamic weighting factor $\alpha$ to control the fusion ratio between the VI estimated distribution and the reconstructed distribution.
It is defined as follows:
\begin{equation}
  \label{eq:alpha}
\setlength\abovedisplayskip{2pt}
  \setlength\belowdisplayskip{2pt}
  %\footnotesize
  \begin{aligned}
    \alpha_{i} = \lambda + (1 - \lambda) \frac{n_i}{(\sum_{j \in N_i} n_j ) / |N_i|}
  \end{aligned}
\end{equation}
, where \(n_i\) is the number of samples of class $i$, and \(\lambda\) is a hyperparameter that controls the balance between these two distributions.

Finally, the aligned mean of the target class i can be calculated as follows:
\begin{equation}
  \label{eq:mu_u}
\setlength\abovedisplayskip{2pt}
  \setlength\belowdisplayskip{2pt}
  %\footnotesize
  \begin{aligned}
    \mu_i^{\prime} = \alpha_i \mu_i + (1 - \alpha_i) \sum_{j \in N_i} \omega_j \mu_j
  \end{aligned}
\end{equation}
For covariance correction, to preserve the directional information of inter-class differences, we introduce an additional covariance mixture term. The overall update method is formulated as follows:

\begin{equation}
  \label{eq:sigma_u}
\setlength\abovedisplayskip{2pt}
  \setlength\belowdisplayskip{2pt}
  %\footnotesize
  \begin{aligned}
    \Sigma_i^{\prime} = & \alpha_i \Sigma_i + (1 - \alpha_i) \sum_{j \in N_i} \omega_j \Sigma_j \\
    & + \alpha_i (1 - \alpha_i)\sum_{j \in N_i} \omega_j (\mu_i -  \mu_j)(\mu_i - \mu_j)^T
  \end{aligned}
\end{equation}

Combining these together, we propose a novel continual learning method for LCIL, named \textbf{ViRN} (\textbf{V}ariational \textbf{I}nference with \textbf{R}edistribution via \textbf{N}eighbors).
This approach integrates two core innovations: variational inference estimates initial class distributions robustly, and trilateration-based interpolation fuses multi-neighbor information to preserve manifold structures.
This dual mechanism yields more discriminative latent representations that enable unbiased sampling via reparameterization, supporting diverse downstream tasks, including classification and generation tasks.

%This approach consists of two key steps: first, variational inference is applied to estimate the initial distribution of each class; second, a redistribution process is conducted using a trilateration-based interpolation to refine the estimated distributions. 
%By leveraging Bayesian inference, our method significantly enhances the robustness of long-tailed distribution estimation compared to conventional statistical approaches. 
%Moreover, the multi-neighbor fusion strategy ensures that the resulting distribution aligns with the geometric centroid of the head classes, preserving the underlying manifold structure.
%This enables {\proposed} to achieve more distinguishable latent space representations, which can facilitate unbiased sampling via reparameterization, thereby supporting diverse downstream tasks including classification and generation.

\begin{table*}[!htb]
  \centering
  \caption{Performance comparison across multiple long-tailed datasets.}
  \label{tab:main_all}
  \resizebox{.6\textwidth}{!}{\begin{tabular}{@{}l|c|c|c|c|c|c|c|c@{}}
  \toprule
  Methods & SC-LT & AM-LT & ESC-LT & US-LT & C100-LT & TIN-LT & Overall\\ 
\midrule
  iCaRL & 35.07 & 43.91 & 20.36 & 46.53 & 82.80  & 83.26 & 311.93 \\
  DGR & 61.79 & 67.23 & 43.08 & 46.79 & 81.86 & 79.62 & 380.37\\ 
  SLCA & \second{83.14} & \second{96.16} & - & 42.16 & \second{88.05} & \second{85.65} & 395.16 \\
  LAE & 78.68 & 80.64 & \second{74.53} & \second{67.07} & 80.49 & 81.41 & \second{462.82} \\
  \midrule
  {\proposed} & \first{85.20} & \first{97.18} & \first{79.33} & \first{85.96} & \first{89.63} & \first{86.96}  & \first{524.26} \\
  \midrule
  \texttt{UpperBound} & 85.35 & 97.21 & 87.88 & 89.07 & 90.62 &  88.39 & 538.52 \\
 \bottomrule
\end{tabular}
}
\vspace{-3mm}
\end{table*}

\section{Experiments}

\begin{figure*}[htb]
\vspace{-1mm}
\setlength{\abovecaptionskip}{-1pt}
\setlength{\belowcaptionskip}{-10pt}
\centering
\includegraphics[width=0.9\linewidth]{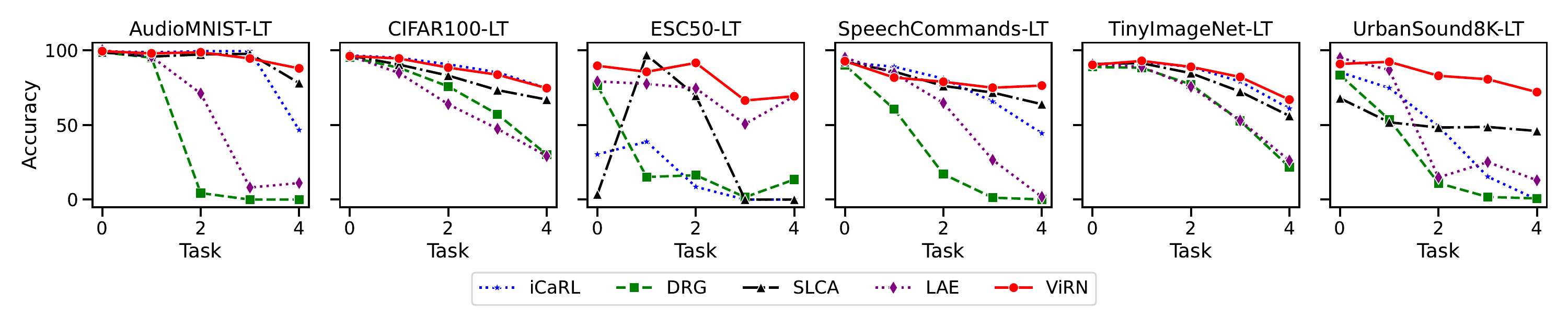}
\vspace{-5mm}
\caption{Accuracy of new class learning over sequential tasks.}\label{fig:num_samples}
\vspace{-3mm}
\end{figure*}

\subsection{Experimental Setup}

\textbf{Datasets.} We evaluated {\proposed} on six long-tailed datasets spanning acoustic and visual domains, with a heavy imbalance ratio ($\rho=0.01$): Speechcommands-LT (SC-LT)~\cite{speechcommandsv2}, AudioMNIST-LT (AM-LT)~\cite{audiomnist2023}, ESC50-LT (ESC-LT)~\cite{piczak2015dataset}, and UrbanSound8K-LT (US-LT)~\cite{Salamon:UrbanSound:ACMMM:14} for acoustic tasks; CIFAR-100-LT (C100-LT)~\cite{Krizhevsky09learningmultiple}, and TinyImageNet-LT (TIN-LT)~\cite{Le2015TinyIV} for image recognition.
All datasets are divided into 5 sequential tasks to simulate incremental learning with long-tailed data streams.

\textbf{Baselines.}
We compared {\proposed} against SOTA class-incremental learning methods:
iCaRL~\cite{Rebuffi2016iCaRLIC}, DGR~\cite{He2024GradientRT}, SLCA~\cite{Zhang2023SLCASL}, and LAE~\cite{Gao2023AUC}. 
An upper-bound model trained on the original, balanced datasets is included to quantify the impact of long-tailed distributions.

\textbf{Implementation.} 
{\proposed} is implemented in PyTorch, leveraging pre-trained backbones:
Wav2Vec~\cite{NEURIPS2020_92d1e1eb} and CLAP~\cite{Wu2022LargeScaleCL} for acoustic embeddings, DINOv2-ViT-B/16~\cite{Oquab2023DINOv2LR} for image features.
All baselines share the same backbone for fairness. Training uses the Adam optimizer ($\text{lr}=1e-5$, $\text{batch-size}=128$) over 100 epochs, with hyperparameters $\lambda=0.7$ (fusion weight) and $k=3$ (nearest neighbors).
%Experiments conducted on an NVIDIA RTX 8000 GPU (48GB).

\subsection{Results}

\textbf{Accuracy.}
We evaluated average Top-1 accuracy across all sequential tasks to measure robustness to catastrophic forgetting and long-tailed bias. As shown in Table~\ref{tab:main_all}, {\proposed} achieves state-of-the-art performance, surpassing prior methods by $61.44\%$. 
Notably, {\proposed} excels on acoustic datasets, which highlights {\proposed}’s ability to handle extreme long-tailed scenarios by leveraging distributional trilateration.
Additionally, {\proposed} narrows the gap with the balanced-data upper bound ($524.26 \text{ vs. } 538.52$), demonstrating slight performance degradation due to long-tailed data. This suggests that our distribution reconstruction mechanism effectively mitigates class imbalance, even in incremental settings.

%This highlights {\proposed}’s ability to handle extreme long-tailed scenarios—even when tail classes contain a single sample—by leveraging distributional trilateration.
%In contrast, methods like SLCA, which rely on statistical replay or feature alignment, collapse on acoustic tasks with severe class imbalance. For example, on ESC50-LT, SLCA’s accuracy drops to $0\%$ by the final task, as its reliance on covariance estimation fails when tail classes lack sufficient samples.

\textbf{Learning of Novel Classes.}
To quantify {\proposed}’s ability to learn new classes under long-tailed data, we analyze its per-task accuracy on classes introduced in each incremental phase (Figure~\ref{fig:num_samples}).
As tasks progress, the number of samples per class diminishes sharply—simulating real-world scenarios where novel categories arrive sparsely. {\proposed} maintains high new-class accuracy across all tasks, achieving a notably high accuracy compared with other baselines in $\task{4}$, where tail-class samples are scarcest. 
This robustness stems from our distribution reconstruction mechanism: by fusing VI-estimated priors with neighborhood-derived distributions, {\proposed} generalizes effectively even from single-sample tail classes.

%While methods like iCaRL exhibit strong initial plasticity, they fail to stabilize learned representations, suffering catastrophic forgetting in later tasks. 
%{\proposed} balances both objectives: its accuracy on newly learned classes remains stable while preserving the best accuracy on overall classes (Table~\ref{tab:main_all}).

\begin{table}[htb]
\vspace{-3mm}
  \centering
  \caption{Ablation study results across multiple datasets (${\uparrow}$ indicates that the higher the better, and vice versa).}
  \label{tab:ablation}
  \resizebox{0.42\textwidth}{!}{\begin{tabular}{@{}l|cc|c|c|c@{}}
  \toprule
  \multirow{1}{*}{Datasets} & VI & RD &  {Accuracy}(\(\uparrow\)) & Forgetting(\(\downarrow\)) & Novelty(\(\uparrow\))\\ \midrule
  \multirow{3}{*}{AM-LT} & \ding{55} & \ding{55} & 88.34 & 0.0 & 66.71 \\
                        & \ding{51} & \ding{55} & 96.22 & 15.71 & 97.37 \\ 
                        & \ding{51} & \ding{51} & 97.18 & 3.58 & 95.58 \\ \midrule

\multirow{3}{*}{C100-LT} & \ding{55} & \ding{55} & 61.93 & 1.10 & 34.14 \\
                        & \ding{51} & \ding{55} & 89.21 & 12.70 & 86.98 \\ 
                        & \ding{51} & \ding{51} & 89.63 & 10.9 & 87.36 \\ \midrule
\multirow{3}{*}{ESC-LT} & \ding{55} & \ding{55} & 23.88 & 30.23 & 35.42 \\
                        & \ding{51} & \ding{55} & 44.20 & 66.27 & 63.50 \\ 
                        & \ding{51} & \ding{51} & 79.33 & 24.41 & 80.38 \\ \midrule
\multirow{3}{*}{SC-LT} & \ding{55} & \ding{55} & 81.35 & 1.78 & 57.27 \\
                        & \ding{51} & \ding{55} & 83.93 & 10.88 & 82.92 \\ 
                        & \ding{51} & \ding{51} & 85.20 & 2.94 & 80.84 \\ \midrule
\multirow{3}{*}{TIN-LT} & \ding{55} & \ding{55} & 58.35 & 0.55 & 32.54 \\
                        & \ding{51} & \ding{55} & 86.42 & 5.70 & 83.61 \\ 
                        & \ding{51} & \ding{51} & 86.96 & 4.10 & 84.12 \\ \midrule
\multirow{3}{*}{US-LT} & \ding{55} & \ding{55} & 73.54 & 1.49 & 45.77 \\
                        & \ding{51} & \ding{55} & 76.12 & 6.67 & 79.34 \\ 
                        & \ding{51} & \ding{51} & 85.96 & 8.51 & 83.62 \\ 

   \bottomrule
\end{tabular}
}
\vspace{-6mm}
\end{table}

\textbf{Ablation study.}
To validate {\proposed}’s design, we ablated its two core components: variational inference (VI) and distribution redistribution (RD). Table~\ref{tab:ablation} compares the full method against ablated variants across datasets, where {\proposed} consistently outperforms the ablated versions across all datasets. Replacing VI with a statistical method leads to biased distribution estimates for tail classes, especially as sample sizes diminish. For instance, on ESC50-LT, it causes a $44.96\%$ drop in new-class accuracy due to overconfident but erroneous covariance estimates. While reduced estimation variance marginally lowers forgetting, the catastrophic collapse in new-class learning renders the statistic method ineffective in long-tailed CIL.

Without RD, the method easily overfits new classes. Specifically, though this improves new-class accuracy by $2.08\%$ on SC-LT, it amplifies forgetting ($7.94\%$), ultimately degrading overall accuracy by $1.27\%$. This underscores RD's role in stabilizing knowledge transfer across tasks.
The ablation results highlight that {\proposed}'s fusion of VI (for robust uncertainty-aware estimation) and RD (for neighborhood-informed alignment) achieves a better trade-off in overall performance.

\section{Conclusion and Discussion}

We introduced {\proposed}, a framework for LCIL that bridges variational inference and distributional geometry. By encoding latent space with VAE, ViRN mitigates bias toward head classes, while Wasserstein-based trilateration reconstructs tail-class representations by leveraging semantically similar neighbors. Experiments across six benchmarks show ViRN’s superiority, with significant gains in speech and audio tasks, underscoring its applicability to real-world scenarios like voice adaptation and rare sound detection.

%\vspace{-4mm}
%We introduced {\proposed}, a framework for long-tailed class-incremental learning (LCIL) that bridges variational inference and distributional geometry. By modeling class-conditional distributions with VI, ViRN mitigates bias toward head classes, while Wasserstein-based trilateration reconstructs tail-class representations by leveraging semantically similar neighbors. Experiments across six benchmarks show ViRN’s superiority, with significant gains in speech and audio tasks, underscoring its applicability to real-world scenarios like adaptive voice interfaces and rare sound detection.

\section*{Acknowledgements}
This work was supported by the Engineering and Physical Sciences Research Council (EPSRC) grant,  MultiTasking and Continual Learning for Audio Sensing Tasks on Resource-Constrained Platforms [EP/X01200X/1].

\bibliography{main}
\bibliographystyle{icml2025}

\end{document}